\def\year{2022}\relax
\documentclass[letterpaper]{article} 
\usepackage{aaai22}  
\usepackage{times}  
\usepackage{helvet}  
\usepackage{courier}  
\usepackage[hyphens]{url}  
\usepackage{graphicx} 
\usepackage{natbib}  
\usepackage{caption} 
\DeclareCaptionStyle{ruled}{labelfont=normalfont,labelsep=colon,strut=off} 
\usepackage{algorithm}
\usepackage{algorithmic}
\usepackage{amsmath}
\usepackage{tabularx}
\usepackage[most]{tcolorbox} 
\usepackage{latexsym}
\usepackage{arydshln}
\usepackage{tikz}
\usepackage{subcaption}
\usepackage{array}
\usepackage{xr}
\usepackage{xcolor}
\newcommand{\answerYes}[1]{\textcolor{blue}{#1}} 
\newcommand{\answerNo}[1]{\textcolor{teal}{#1}}

\usepackage{newfloat}
\usepackage{listings}

\usepackage{svg}
\usepackage{booktabs}

\usepackage{enumitem}
\usepackage{tcolorbox}
\usepackage{enumitem}
\usepackage{pifont}
\usepackage{setspace}

\usepackage{multirow}
\usepackage{longtable}

\usepackage{algorithm}
\usepackage{algorithmic}

\usepackage{newfloat}
\usepackage{listings}
\DeclareCaptionStyle{ruled}{labelfont=normalfont,labelsep=colon,strut=off} 
\floatstyle{ruled}
\newfloat{listing}{tb}{lst}{}
\floatname{listing}{Listing}
\usepackage{svg}
\usepackage{booktabs}
\usepackage{enumitem}

\usepackage{subcaption}
\usepackage{caption}
\usepackage{color}
\usepackage{xcolor}
\usepackage{soul}
\usepackage{longtable}
\usepackage{multirow}
\usepackage[draft,textsize=footnotesize,textwidth=15mm]{todonotes}

\title{PHAnToM: Persona-based Prompting Has An Effect on Theory-of-Mind Reasoning in Large Language Models}

\author{Fiona Anting Tan$^{1}$, Gerard Christopher Yeo$^{1}$, \textbf{Kokil Jaidka}$^{2}$, {Fanyou Wu}$^{3}$, \textbf{Weijie Xu}$^{3}$, \\ \textbf{Vinija Jain}$^{3,4}$, \textbf{Aman Chadha}$^{3,4}$, \textbf{Yang Liu}$^{5}$, \textbf{See-Kiong Ng}$^{1}$}

\affiliations {
    \textsuperscript{\rm 1}Institute of Data Science\\
    \textsuperscript{\rm 2}NUS Centre for Trusted Internet and Community, National University of Singapore \\
    \textsuperscript{\rm 3}Amazon\\
    \textsuperscript{\rm 4}Stanford University\\
    \textsuperscript{\rm 5}Tsinghua University\\
    \texttt{\{tan.f, e0545159\}@u.nus.edu} 
}

\usepackage{bibentry}

\begin{document}

\maketitle

\begin{abstract}
The use of LLMs in natural language reasoning has shown mixed results, sometimes rivaling or even surpassing human performance in simpler classification tasks while struggling with social-cognitive reasoning, a domain where humans naturally excel. These differences have been attributed to many factors, such as variations in prompting and the specific LLMs used. However, no reasons appear conclusive, and no clear mechanisms have been established in prior work. In this study, we empirically evaluate how role-playing prompting influences Theory-of-Mind (ToM) reasoning capabilities.  Grounding our rsearch in psychological theory, we propose the mechanism that, beyond the inherent variance in the complexity of reasoning tasks, performance differences arise because of socially-motivated prompting differences. In an era where prompt engineering with role-play is a typical approach to adapt LLMs to new contexts, our research advocates caution as models that adopt specific personas might potentially result in errors in social-cognitive reasoning.
\end{abstract}

%

\section{Introduction}

\begin{figure}[!h]
\centering
  \includegraphics[scale=0.25]{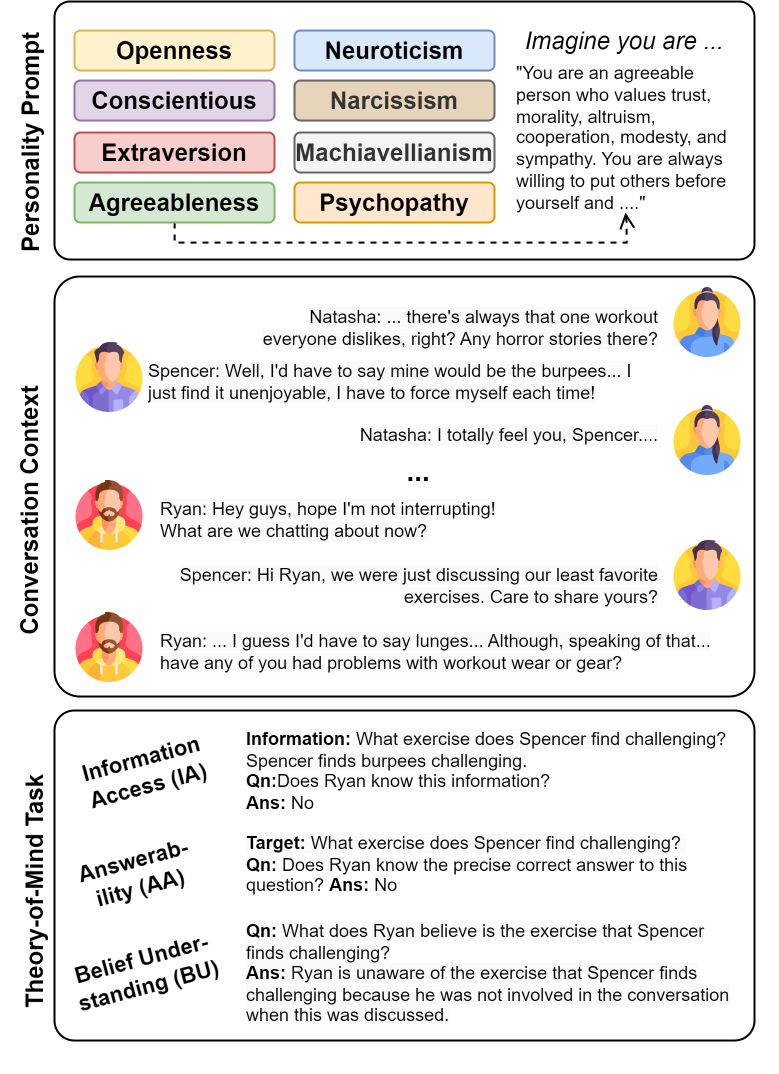}
  \caption{Overview of PHAnToM. Our work investigates how eight different persona-based prompts (Big Five OCEAN and Dark Triad) affects LLMs' ability to perform three theory-of-mind reasoning tasks (Information Access (IA), Answerability (AA), and Belief Understanding (BU)).}
  \label{fig:introduction}
\end{figure}
Large language models (LLMs) have demonstrated impressive capabilities across a variety of natural language processing (NLP) tasks \cite{lyu2023new, bai2023benchmarking, bang2023multitask}. However, these models have been reported to exhibit generally inadequate social-cognitive reasoning abilities \cite{farha2022semeval, perez2022semeval}, which are crucial for applications involving human interaction. One particularly important social-cognitive reasoning task is the Theory-of-Mind (ToM) task \cite{kosinski2023theory, premack1978does}, traditionally studied in the context of human development. ToM refers to the ability to attribute mental states—such as beliefs, intentions, thoughts, and emotions—to oneself and others, a capability essential for effective communication and interaction \cite{gallese2011so, wimmer1983beliefs}.

While some studies suggest that LLMs display a degree of ToM abilities \cite{kim-etal-2023-fantom, ma2023tomchallenges, shapira-etal-2023-well}, these models remain significantly inferior to humans in this domain. This discrepancy between human and LLM performance in ToM tasks presents a challenge, especially as LLMs are increasingly deployed in settings that require sophisticated human interaction. The inadequacy of LLMs in ToM tasks motivates the need to explore strategies that could enhance their social-cognitive reasoning capabilities. Furthermore, although there has been progress in assessing both ToM abilities and role-oriented prompt engineering in LLMs, these areas of research have largely been studied in isolation. A second research gap lies in the lack of explanatory mechanisms offered in prior studies for why different styles of prompting might lead to varying levels of ToM performance.

Our study advocates performance audits with persona-based prompting: a technique that uses personality traits to characterize personas with distinct social and cognitive motivations—and evaluates their effect on ToM reasoning abilities. This approach is motivated first by recent computational linguistics research that recognizes the psychological dimensions underlying interpersonal conversations, which we have adapted and applied to the Human-AI instruction context~\cite{liu2023psyam,dutt2020keeping,giorgi2024findings}. Second, we draw from prior psychological research that links personality dimensions to social-cognitive reasoning, and suggests that personality traits influence ToM abilities in humans~\cite{mccrae1992introduction,john2008paradigm}. Third, recent NLP research has demonstrated that persona-based prompting provides reasonable adherence to responses by synthetic humans~\cite{rathje2024gpt}. Accordingly, we report experiments that examine how different personality traits affect ToM abilities in LLMs. Our approach comprises experimenting with persona-based prompting techniques to induce specific personality traits for solving three ToM tasks with three different LLMs, specifically GPT-3.5, Llama 2, and Mistral. Our analyses evaluated the effects of the persona-based prompts on the LLMs' performance across the three tasks. Our research answers three key questions:
\begin{itemize}
    \item How does persona-based prompting influence model performance in ToM tasks?
    \item Which LLMs exhibit the highest and lowest sensitivity to persona-based prompting across different tasks?
    \item How do the cumulative effects of persona-based prompting influence model performance in ToM tasks?
\end{itemize}
Based on the insights, we propose a mechanism to explain why different persona-based prompting styles elicit varying levels of ToM performance. Our theoretical contributions include:
\begin{itemize}
    \item Providing novel evidence that Persona-based Prompting Has An Effect on Theory-of-Mind (PHAnToM) reasoning in LLMs, with Dark Triad traits having a larger impact than Big Five traits on ToM performance across models and tasks.
    \item Demonstrating that LLMs with higher variance across persona-based prompts in ToM tasks tend to be more controllable in personality tests.
    \item Contextualizing our observations about ToM abilities in LLMs within the broader framework of psychological theories on human cognition.
\end{itemize}
In a landscape where role-play is increasingly common in LLM applications, our study is the first to explore the intersection of persona-based prompting and ToM abilities in LLMs. Our findings suggest that persona-based prompting, particularly when aligned with specific personality traits, can influence ToM task performance in predictable ways. This also highlights the importance of carefully considering the personas assigned to LLMs, as these can significantly shape their reasoning abilities based on inferred social and cognitive motivations.

\section{Related Work}
\label{sec:related}
\subsection{Sensitivity of LLMs to Prompts}

Multiple research studies have shown the brittleness of LLMs to the input prompts. Zero shot Chain-of-Thought (incorporating one-line in prompts, like ``\emph{First,}'' or ``\emph{Let's think step by step}'') \cite{DBLP:conf/nips/KojimaGRMI22, bsharat2023principled} has empirically allowed LLMs to become stronger reasoners, especially for arithmetic tasks. In other works, strategies like role-play (including a description of someone the LLM should embody) \cite{DBLP:journals/corr/abs-2308-07702} or threats (reminding the LLM they would be penalized if they answer wrongly, or that the users' life matters gravely on this answer) \cite{bsharat2023principled} have also demonstrated effectiveness in improving LLM performance. \citet{sclar2023quantifying} find that small prompt variations often yield large performance differences. \citet{wu2023language} showed that with Instruction Fine-tuning, LLMs can distinguish instruction with context and focus more on instructions. They further show that instruction fine-tuning encourages self-attention heads to encode more word-word relations related to instruction verbs. \citet{gupta2024bias} found that LLM's reasoning abilities can be affected by persona prompts across different socio-demographic groups (race, gender, religion, disability, and political affiliation). Encouraged by these findings, we were inspired to examine the sensitivities of LLMs to personality role-play via prompting on socio-cognitive reasoning in LLMs.

\subsection{Inducing Personas in LLMs with Prompts}
Personality refers to the enduring and stable characteristic patterns of cognitions, feelings, and behaviors, generally consistent across situations \cite{allport1937personality}. A persona, in this context, is a constructed identity or role that an LLM adopts, which is shaped by specific personality traits. While we induce personas through descriptions of personality traits, the use of a different term implies our acknowledgment that the two are not pseudonymous.

Prior work on personality has primarily applied the five-factor model (or Big Five) of personality \cite{john2008paradigm} as the framework of choice to analyze individual differences. It comprises five subscales: openness, conscientiousness, extraversion, agreeableness, and neuroticism traits (OCEAN) \cite{mccrae1992introduction}. Psychometric tests such as the International Personality Item Pool (IPIP-NEO) \cite{goldberg1999broad}, and the Big Five Inventory (BFI) \cite{john1999big} are commonly used to measure these traits in humans. 

Recently, \citet{DBLP:journals/corr/abs-2206-07550, DBLP:journals/corr/abs-2307-00184, DBLP:journals/corr/abs-2312-14202} administered these psychometric tests on LLMs under specific prompting configurations and found that it is possible to obtain reliable and valid personality measurements with LLMs, implying that the prompts successfully induced personas that align with the intended personality traits. Furthermore, by introducing role-play prompts, they demonstrated the adaptability of LLMs, where personalities can be shaped along desired dimensions to simulate specific human personality profiles. These results could be explained by psycholinguistic studies that showed certain expressed linguistic features reliably reflect personality traits \cite{boyd2017language}. However, these studies do not account for the potential slippage or variability in task performance that may arise when LLMs adopt these induced personas. Our work adapts their approaches and extends their work to examine how these personas influence LLM performance on specific tasks, particularly in the context of Theory-of-Mind reasoning.

\subsection{Theory-of-Mind Reasoning}
ToM is typically assessed using the false belief paradigm \cite{beaudoin2020systematic, wellman2001meta, wimmer1983beliefs}, with the ``Sally and Ann" task being a prototypical example \cite{baron1985does}. In this task, humans typically succeed between 3 and 5 years of age \cite{wellman2001meta}, as they develop the understanding that different agents can hold different beliefs about the world, and that these beliefs may be inconsistent with reality. ToM is crucial for effective social communication, adaptation, and forming higher quality social relationships \cite{fink2015friends, imuta2016theory}, as it allows individuals to infer the beliefs, desires, and intentions of others and to act accordingly in various situational contexts.

Recent works have explored LLMs' ToM abilities across a variety of tasks \cite{kim-etal-2023-fantom, ma2023tomchallenges, shapira-etal-2023-well}. Generally, the results suggest that while LLMs exhibit some degree of ToM, their performance still lags behind that of humans. For instance, when presented with a narrative or full conversation as a prompt, LLMs often adopt an omniscient-view belief in ToM tasks, evaluating all of the information provided and producing incorrect outputs without recognizing that certain agents did not possess the same belief \cite{kim-etal-2023-fantom}.

Despite these insights, there are several research gaps that need to be addressed to better understand the complexities of ToM in LLMs. First, current studies often focus on a single type of reasoning task, such as belief attribution, without considering how different facets of reasoning might be influenced by varying cognitive demands and task complexities. Second, the influence of induced personas on LLMs' ToM reasoning across tasks of varying complexity and cognitive demands remains underexplored. To address these gaps, we employ an experiment design that considers the interaction between persona-based prompting methods and the complexity of ToM tasks.


\section{Methodology}
\label{sec:method}
Figure \ref{fig:introduction} outlines the key investigations explored in this work. In summary, we examined the effects of persona-based prompting on ToM reasoning capabilities in LLMs. Our investigations covered eight personas and three ToM tasks.

\subsection{Prompting strategies}
Persona-based prompting was conducted through a set of eight personality traits. The description and prompt for each personality trait were designed based on theoretical formulations of the trait in the personality psychology literature and informed by validated psychometric measures \cite{gosling2003very,jonason2010dirty, jones2014introducing, mccrae1987validation}. One of our authors, a psychology graduate with training in personality psychology, reviewed the wording and phrasing of the descriptions to ensure they were appropriate for input into the LLMs. The actual descriptions of the persona-based prompts can be found in the Supplementary Materials.

\paragraph{The Big Five OCEAN}:
\begin{itemize}
    \item \textbf{Openness:} Reflects the extent to which a person is open to new experiences and ideas. Individuals with high scores tend to be curious, imaginative, and open-minded, while those with low scores may prefer routine and familiarity.
    \item \textbf{Conscientiousness:} Reflects the degree of organization, responsibility, and reliability in a person. High scorers are often diligent, organized, and goal-oriented, while low scorers may be more spontaneous and less focused on long-term planning.
    \item \textbf{Extraversion:} Reflects the level of sociability, assertiveness, and energy a person exhibits. High scorers are typically outgoing, energetic, and enjoy social interactions, whereas low scorers may be more reserved and introverted.
    \item \textbf{Agreeableness:} Reflects interpersonal relations and cooperation. Individuals with high agreeableness scores are often compassionate, cooperative, and considerate, while low scorers may be more competitive or assertive.
    \item \textbf{Neuroticism:} Reflects emotional stability and reaction to stress. High scores indicate emotional instability, anxiety, and moodiness, while low scores suggest emotional resilience and a more stable emotional state.    
\end{itemize}

\paragraph{The Dark Triad}:
\begin{itemize}
    \item \textbf{Narcissism:} Reflects a sense of entitlement, superiority to others, and grandiosity. Moreover, narcissists like to be the center of attention, associate with famous or popular people, and often display an arrogant demeanor towards others.
    \item \textbf{Machiavellianism:} Reflects interpersonal coldness towards others and a tendency to manipulate and exploit others through deception and flattery to achieve one's goals. Individuals high in this trait plan and act primarily for their own benefit. 
    \item \textbf{Psychopathy:} Reflects low empathy towards others and a tendency to exhibit thrill-seeking behaviors without concern for negative moral consequences. Individuals high in this trait lack remorse, often seek revenge on others, and especially target authorities.
\end{itemize}

\begin{figure*}[!ht]
  \hspace*{-2mm}\includegraphics[scale=0.25]{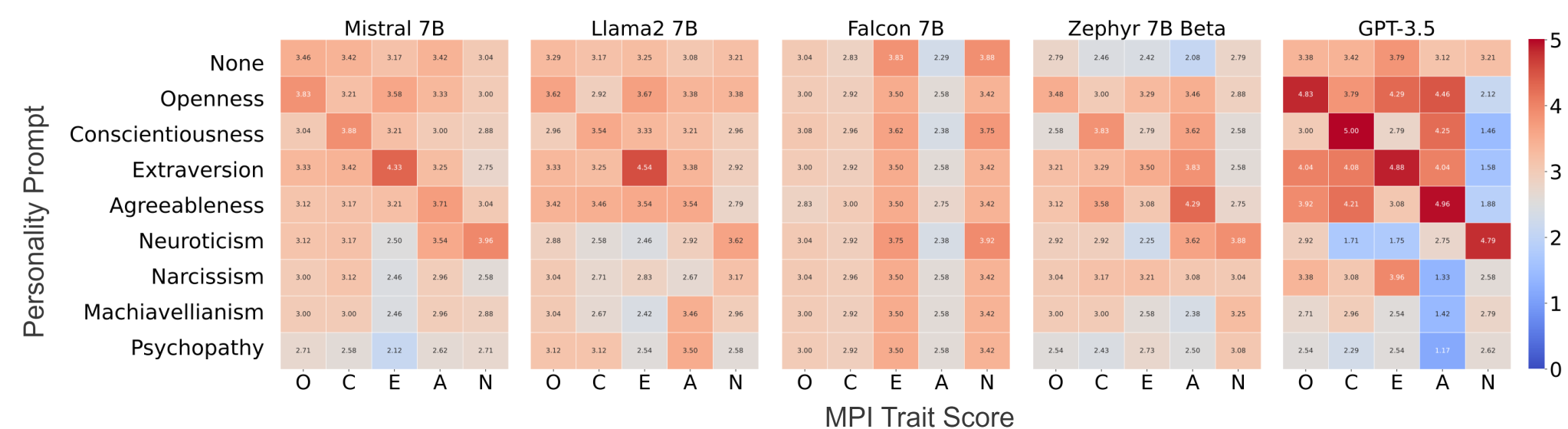}
  \vspace*{-4mm}
  \caption{Heatmap of MPI120 scores for the Big Five OCEAN traits (x-axis) when models are prompted with different personalities (y-axis). Scores range from 0 (Blue) to 5 (Red).}
  \label{fig:mpi_scores}
\end{figure*}
\subsection{Theory-of-Mind Reasoning Task}
The following paragraphs detail how the three ToM tasks from the FANTOM dataset \cite{kim-etal-2023-fantom} are operationalized:
\begin{itemize}
\item \textbf{Information Access (IA):} A binary classification task where models determine if a character has knowledge or access to certain information based on their presence in a conversation. This task assesses whether a character who was absent during part of the conversation has the same knowledge as those who were present \cite{wellman2011sequential, wellman2018theory}.
    \item \textbf{Answerability (AA):} A binary classification task that extends IA by requiring models to determine not just access to information but also whether a character can answer a question correctly. This task evaluates reasoning about a character’s ability to respond based on the information they possess.

\item \textbf{Belief Understanding (BU):} A multiple-choice task requiring models to infer the beliefs of characters. This is the most challenging task \cite{wellman2018theory}, as models must recognize differing beliefs between characters, assess information access, and identify false beliefs, even when the model knows the correct answer.

\end{itemize}

The data sizes are 3571 for IA and AA each, and 993 for BU. 

\subsection{Triggering Personalities in LLMs for ToM}

We followed typical LLM role-play procedures by including the prefix ``\emph{Imagine you are someone that fits this description: \{personality\_description\}}'' prepended to the context and task question itself. 



For the purpose of better clarifying the mechanisms underlying model performance, the personas were ordered from least to most social based on their established relationships with social behavior and interpersonal interactions as identified in the literature. Psychopathy was placed first as it reflects low empathy and antisocial tendencies, making it the least social \cite{jonason2010dirty, jones2014introducing}. Machiavellianism follows, as it involves manipulative and self-serving behavior that lacks genuine social concern \cite{jonason2010dirty, jones2014introducing}. Narcissism is next, characterized by a need for admiration and attention, which, while involving social interaction, is still primarily self-focused \cite{jonason2010dirty, jones2014introducing}. The Big Five traits of Neuroticism, Openness, Conscientiousness, Extraversion, and Agreeableness were then ordered, with Agreeableness being the most social, reflecting empathy, cooperation, and concern for others \cite{mccrae1992introduction, john2008paradigm, baron2004empathy}.


\subsection{Experimental setup}
\label{sec:findings}
We explored an array of state-of-the-art LLMs, namely \textbf{Mistral 7B} (\texttt{Mistral-7B-Instruct-v0.1}) \cite{DBLP:journals/corr/abs-2310-06825}, \textbf{Llama 2} (\texttt{Llama-2-7b-chat-hf}) \cite{DBLP:journals/corr/abs-2307-09288}, \textbf{Falcon 7B} (\texttt{falcon-7b-instruct}) \cite{falcon40b}, \textbf{Zephyr 7B Beta} (\texttt{zephyr-7b-beta}) \cite{DBLP:journals/corr/abs-2310-16944}, and OpenAI \textbf{GPT-3.5} (\texttt{gpt-3.5-turbo-1106}). We worked with the Instruct versions of the models, which were designed to respond to tasks, instead of the vanilla versions of the models, for better performance. Similar to \citet{kim-etal-2023-fantom}, we report weighted F1 scores for IA and AA, and accuracy for BU. We apply a random seed of 99 for all experiments. For all models available on Huggingface Hub (all except GPT-3.5), greedy decoding was used. More details about the models' hyperparameters are in the Supplementary Materials.


\section{Results}
\subsection{Manipulation Checks of Persona-based Prompts}
\label{sec:mpi}
Figure \ref{fig:mpi_scores} presents the results from our manipulation check of the prompts. It shows a heatmap of MPI120 scores for the Big Five OCEAN traits across different models (Mistral 7B, Llama 2, Falcon 7B, Zephyr 7B Beta, and GPT-3.5) when prompted with different personality traits. Scores range from 0 (Blue) to 5 (Red). When these models were prompted with specific target personality traits, we would anticipate a significant increase in the corresponding scores of the target personality traits on the MPI. GPT-3.5 exhibited the highest correspondence to the prompted traits, particularly showing stronger alignment with traits like Conscientiousness and Agreeableness, while other models display more moderate and consistent responses across all traits. Falcon's MPI scores remained consistent across persona-based prompts, suggesting robustness to such prompts or that it has been previously trained on data that instructs them to ignore potentially malicious instructions.

The procedure followed was similar to \citet{DBLP:journals/corr/abs-2206-07550}. We first administered the Machine Personality Inventory (MPI) on the LLMs to check whether the persona-based prompting successfully simulates the respective traits. The MPI consists of 120 questions adapted from various psychometrically valid personality scales, measuring the Big Five OCEAN personality traits. Each question presents a statement of a trait (e.g., ``\emph{You have difficulty imagining things}'') and the LLM is tasked to rate the accuracy of how this statement describes them on a 5-point Likert scale.\footnote{The 5-point options available were: (A). Very Accurate, (B). Moderately Accurate, (C). Neither Accurate Nor Inaccurate, (D). Moderately Inaccurate, and (E). Very Inaccurate}

\begin{figure}[!ht]
\centering
\includegraphics[width=0.43\textwidth, scale=0.20]{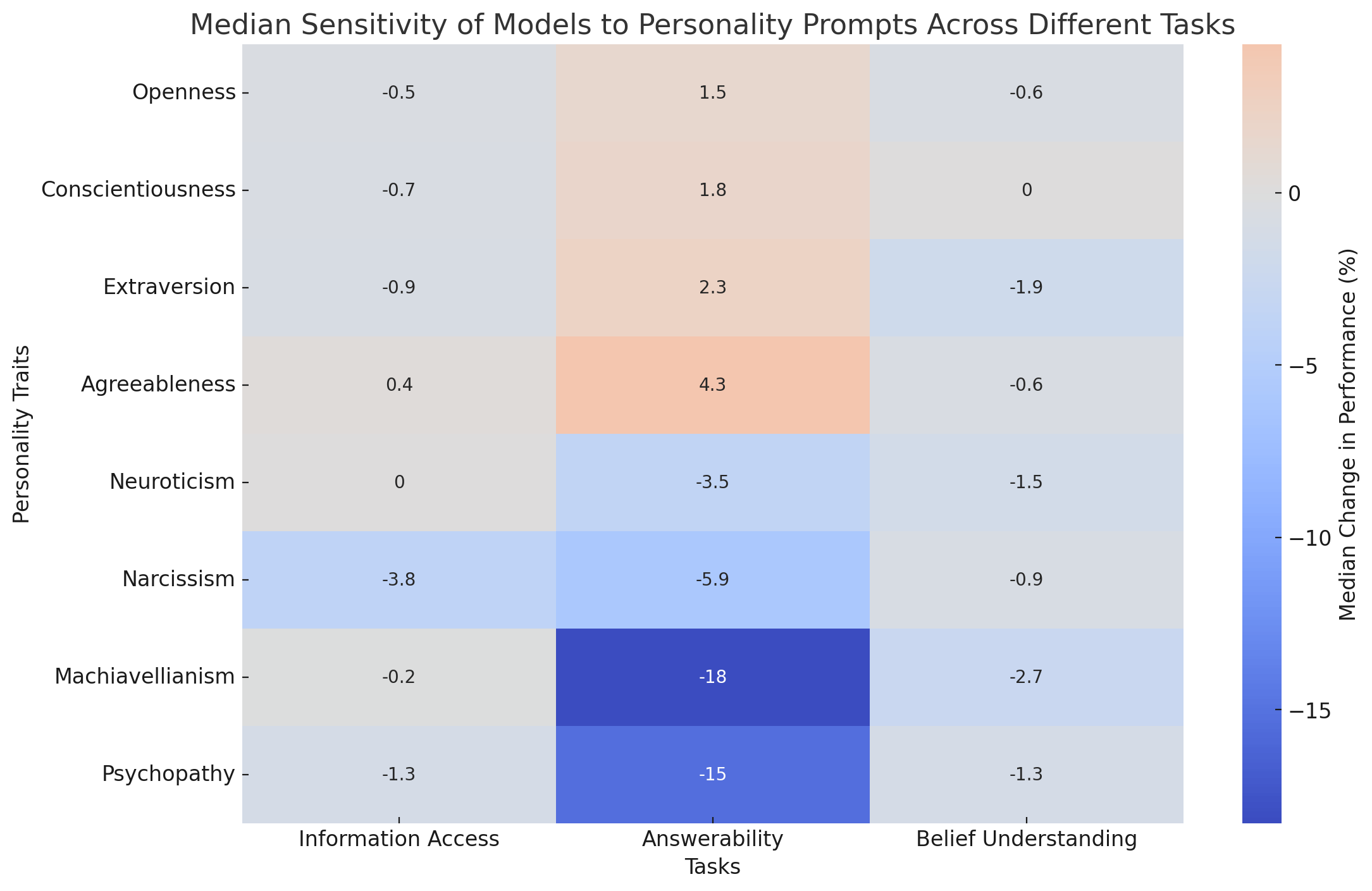} 
\caption{Median performance change across models, when compared to models' baseline performances without persona-based prompting.}
\label{fig:rq1}
\end{figure}
\begin{table*}[ht!]
\small

\begin{subtable}[]{\textwidth}
\small
\caption{Information Access Task}\label{tab:info_access}
\centering
\resizebox{0.65\columnwidth}{!}{
\begin{tabular}{cccccc}
\toprule
\textbf{Personality} & \textbf{Mistral 7B} & \textbf{Llama 2} & \textbf{Falcon 7B} & \textbf{Zephyr 7B Beta} & \textbf{GPT-3.5} \\
\toprule
\emph{None} & 71.5 & \ul{45.4} & 62.4 & \textbf{40.3} & 59.8 \\
\midrule
Openness & -0.5 & +2.1 & \ul{-0.2} & -2.9 & +0.9 \\
Conscientious & -0.7 & \color{blue}{\textbf{+8.3}} & 0.0 & \ul{-4.5} & +1.1 \\
Extraversion & -0.9 & +2.1 & +0.4 & -3.1 & +0.6 \\
Agreeableness & -0.3 & +3.1 & +0.5 & -3.8 & +1.0 \\
Neuroticism & \textbf{0.0} & +4.2 & +0.1 & -2.1 & +1.9 \\
Narcissism & -0.3 & \color{blue}{+6.9} & +1.3 & -1.3 & +4.5 \\
Machiavellianism & -0.2 & \color{blue}{+6.3} & \textbf{+0.9} & -0.6 & \color{blue}{\textbf{+5.0}} \\
Psychopathy & \ul{-1.3} & \color{blue}{+7.0} & +0.7 & -1.4 & +1.8\\
\bottomrule
\end{tabular}
}
\end{subtable}


\begin{subtable}[]{\textwidth}
\small
\caption{Answerability Task}\label{tab:answerability}
\centering
\resizebox{0.65\columnwidth}{!}{
\begin{tabular}{cccccc}
\toprule
\textbf{Personality} & \textbf{Mistral 7B} & \textbf{Llama 2} & \textbf{Falcon 7B} & \textbf{Zephyr 7B Beta} & \textbf{GPT-3.5} \\
\toprule
\emph{None} & 54.1 & 54.6 & \ul{44.5} & 50.7 & \textbf{61.9} \\
\midrule
Openness & +1.5 & +1.8 & +1.1 & 0.0 & -2.8 \\
Conscientious & \color{blue}{\textbf{+6.6}} & \textbf{+3.8} & +0.6 & \ul{-1.2} & \color{red}{\ul{-6.7}} \\
Extraversion & +2.3 & +1.5 & +1.2 & -0.3 & -2.7 \\
Agreeableness & +4.3 & \textbf{+3.8} & +1.4 & +0.2 & -4.3 \\
Neuroticism & +1.3 & \color{red}{-8.0} & +0.1 & -0.3 & -1.7 \\
Narcissism & -2.6 & \color{red}{-21.8} & +0.3 & +1.6 & -3.7 \\
Machiavellianism & \color{red}{\ul{-9.1}} & \color{red}{\ul{-33.0}} & \textbf{+1.8} & \textbf{+1.7} & -3.1 \\
Psychopathy & \color{red}{-8.1} & \color{red}{-27.8} & \ul{-0.3} & +1.0 & -3.2\\
\bottomrule
\end{tabular}
}
\end{subtable}


\begin{subtable}[]{\textwidth}
\small
\caption{Belief Understanding Task}\label{tab:belief_understanding}
\centering
\resizebox{0.65\columnwidth}{!}{
\begin{tabular}{cccccc}
\toprule
\textbf{Personality} & \textbf{Mistral 7B} & \textbf{Llama 2} & \textbf{Falcon 7B} & \textbf{Zephyr 7B Beta} & \textbf{GPT-3.5} \\
\toprule
\emph{None} & 16.1 & 16.0 & 47.5 & 21.5 & \ul{9.6} \\
\midrule
Openness & -0.6 & \ul{-0.5} & \textbf{+0.1} & -0.7 & +0.9 \\
Conscientious & -0.3 & +0.4 & 0.0 & -0.4 & +0.1 \\
Extraversion & \ul{-1.9} & -0.3 & 0.0 & -0.5 & +1.3 \\
Agreeableness & +0.6 & -0.4 & 0.0 & \ul{-1.4} & +0.2 \\
Neuroticism & +0.6 & +0.4 & 0.0 & -0.9 & +0.6 \\
Narcissism & -0.2 & +1.2 & 0.0 & -0.9 & +1.9 \\
Machiavellianism & +2.6 & \textbf{+4.1} & 0.0 & 0.0 & \color{blue}\color{blue}{\textbf{+6.2}} \\
Psychopathy & \textbf{+3.8} & +2.5 & 0.0 & \textbf{+0.5} & +2.0 \\
\bottomrule
\end{tabular}
}
\end{subtable}

\caption{Weighted F1 scores IA and AA, and Accuracy for BU across models and personality prompts. For each model and task, we show the change in scores against the models' performance without any personality prompt. Highest (Lowest) score per column is \textbf{bolded} (\ul{underlined}). Scores that increase (decrease) by 5 or more points are colored \color{blue}{blue} \color{black}{(}\color{red}{red}\color{black}{)}.}\label{tab:tom_scores}
\end{table*}
\subsection{Main Findings}
\label{sec:findings}
Our first research question asks, \textbf{how does persona-based prompting influence model performance in ToM tasks?} While the detailed results are reported in Table~\ref{tab:tom_scores}, Figure \ref{fig:rq1} illustrates the median sensitivity of models to persona-based prompts across different Theory of Mind (ToM) tasks, covering Information Access, Answerability, and Belief Understanding tasks. Each cell in the heatmap shows the median change in performance (in percentage) for a given personality trait and task. The color gradient reflects the magnitude of change, with blue representing performance declines and red/orange showing improvements. The scale on the right indicates the degree of performance shift.

We observe that persona-based prompts do affect the LLMs' performance on ToM tasks. This effect is most pronounced for the Answerability task. For instance, Machiavellianism caused a significant drop in performance, with a change of -18\% and -33.0 (Llama 2), and -27.8 (Psychopathy in Llama 2) in the Answerability task. Similarly, in the Belief Understanding task, Machiavellianism results in -2.7\% change. In contrast, for models like GPT-3.5, the highest positive shift came from Agreeableness in the Answerability task (4.3\%) and Psychopathy in the Belief Understanding task (+4.1).

Across all tasks, the Dark Triad traits (Machiavellianism, Narcissism, and Psychopathy) and Neuroticism are associated with adverse effects on ToM reasoning. In contrast, other personality traits, such as Agreeableness and Conscientiousness, are associated with improved ToM reasoning scores. Notably, GPT-3.5 showed resilience with minor variations, including increases with Agreeableness (+5.0) in the Information Access task. Overall, the Dark Triad traits tend to cause sharper performance shifts compared to the Big Five OCEAN traits, particularly in Answerability tasks.

Table~\ref{tab:tom_scores} breaks down the specific impact of personality traits on model performance across different tasks for four models: Mistral 7B, Llama 2, Falcon 7B, and GPT-3.5. The table highlights changes in performance scores, measured as the weighted F1 score for Information Access (IA) and Answerability (AA) tasks, and accuracy for Belief Understanding (BU). For Information Access, the baseline (no prompt) shows that GPT-3.5 outperforms the other models, achieving a score of 59.8. Conscientiousness has the most positive impact on Llama 2 (+8.3), while Machiavellianism significantly boosts performance in Mistral 7B (+6.3) and Falcon 7B (+0.9), showing the varying sensitivity of models.

In the Answerability task, we observe a much wider range of performance changes. Machiavellianism and Psychopathy sharply reduce performance in Llama 2 by -33.0 and -27.8 points, respectively. However, Agreeableness enhances GPT-3.5's performance (+4.3) and also benefits Mistral 7B (+3.8). These stark variations across models and traits suggest that prompting for ToM tasks requiring reasoning and contextual understanding can yield different results even if models are essentially built on the same architecture, such as Llama.

For the Belief Understanding task, while models generally struggle more, we observe a similarly negative effect from Dark Triad traits like Psychopathy (-2.7) and Machiavellianism (-2.7) in Falcon 7B, though Mistral 7B sees minor improvements (+2.6) from Machiavellianism. Across all tasks, GPT-3.5 shows resilience, with moderate variations across traits but no extreme dips, indicating more balanced performance across traits compared to other models.

The differences in the influences of persona-based prompting on various tasks could be attributed to the specific ToM constructs assessed by each task. Different ToM tasks are proposed to measure different facets of ToM which might not be correlated to one another \cite{nettle2008agreeableness, wellman2001meta}. The Belief Understanding task assesses false beliefs where models need to determine whether a character absent from parts of the conversation will make a false belief about the information discussed, instead of just inferring whether a character has access to particular information (Information Access). From our results, the Answerability task is most susceptible to persona-based prompts, suggesting that it has a greater influence on the models’ abilities to determine the correct answer in addition to understanding about characters’ access to information. Overall, the findings shed light on our first research question on how different psychological traits have differential effects on various ToM tasks. 

\begin{figure}[!ht]
\centering
\includegraphics[width=0.5\textwidth, scale=0.2]{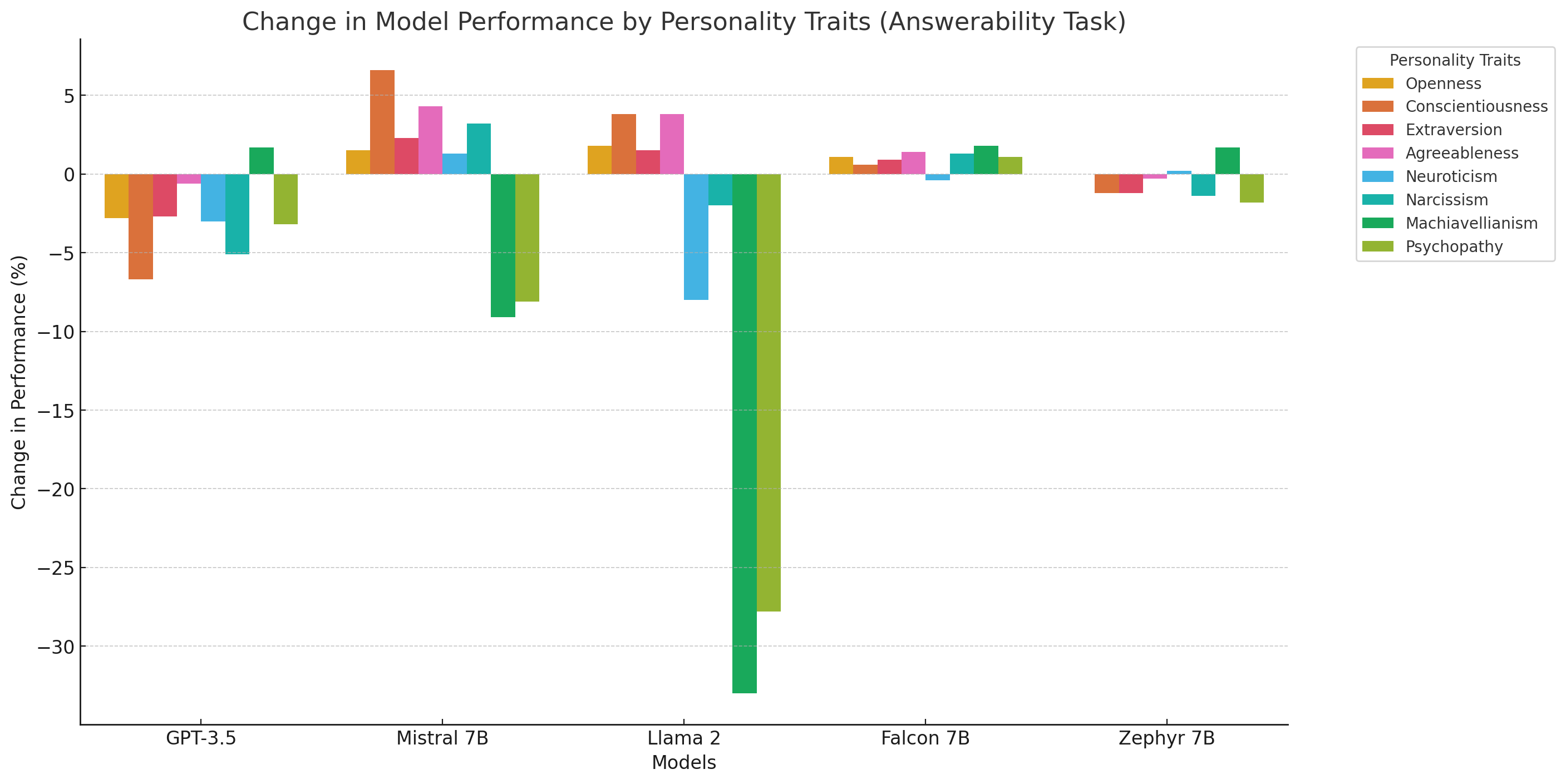} 
\caption{Sensitivity of models to persona-based prompts for Answerability Task.}
\label{fig:rq2}
\end{figure}

Our second research question asks, \textbf{which LLMs exhibit the highest and lowest sensitivity to
persona-based prompting across different tasks?} Based on the prior finding, in Figure \ref{fig:rq2} we focus on the models' sensitivity to persona-based prompts for the Answerability scores. The results for Information Access and Belief Understanding are found in the Supplementary Materials. 

Figure \ref{fig:rq2} offers a detailed comparison of how each model's performance changes in response to different persona-based prompts during the Answerability task, where each bar in the bar chart represents the raw performance change for a specific personality trait, with different colors representing different personality traits. For example, Llama 2 shows a significant negative impact when prompted with Machiavellianism and Psychopathy, whereas Mistral 7B shows improvements with Conscientiousness and Extraversion.

Across models, Llama 2 demonstrates the highest sensitivity to persona-based prompts, with a notable 33\% decrease in F1 score, followed by Mistral (9.1\%), GPT-3.5 (6.7\%), and Zephyr (4.5\%). In contrast, Falcon is observed to be relatively resistant to persona-based prompts, with a shift of at most 1.8\%. This variation may stem from differences in model training methodologies. Llama 2 and GPT-3.5 were fine-tuned using Reinforcement Learning with Human Feedback (RLHF). Studies have shown that RLHF models are more sensitive to personality descriptions \cite{DBLP:journals/corr/abs-2307-00184} and also obtain personality scores that are more aligned with humans \cite{DBLP:journals/corr/abs-2206-07550}. Although Mistral did not undergo RLHF, its training data from publicly available instruction datasets on Huggingface likely containing human-generated content contributes to its sensitivity to persona-based prompts. Conversely, Zephyr and Falcon were predominantly fine-tuned on LLM-generated dialogues, and therefore, possibly exhibit lower sensitivity due to their limited exposure to personality-based questions or terms during fine-tuning. Overall, the findings show that different models have different sensitivities to ToM scores by persona-based prompting, addressing our second research question. 

\begin{table*}[ht!]
\centering
\begin{tabular}{lrrclrrclrrc}
\toprule
\multicolumn{1}{l}{} & \multicolumn{3}{c}{\textbf{(A) Information Access}} & \multicolumn{1}{l}{} & \multicolumn{3}{c}{\textbf{(B) Answerability}} & \multicolumn{1}{l}{} & \multicolumn{3}{c}{\textbf{(C) Belief Understanding}} \\ 
\cmidrule[0.75pt]{1-4} \cmidrule[0.75pt]{6-8} \cmidrule[0.75pt]{10-12} 

\multicolumn{1}{l}{} & \multicolumn{1}{l}{Ours} & \multicolumn{1}{l}{Alt} & \multicolumn{1}{c}{$\Delta$} & \multicolumn{1}{l}{} & \multicolumn{1}{l}{Ours} & \multicolumn{1}{l}{Alt} & \multicolumn{1}{c}{$\Delta$} & \multicolumn{1}{l}{} & \multicolumn{1}{l}{Ours} & \multicolumn{1}{l}{Alt} & \multicolumn{1}{c}{$\Delta$} \\ \cmidrule[0.75pt]{2-4} \cmidrule[0.75pt]{6-8} \cmidrule[0.75pt]{10-12}

Openness & 71.0 & 70.8 & -0.2 &  & 55.6 & 55.0 & -0.6 &  & 14.5 & 14.9 & +0.4 \\
Conscientious & 70.8 & 70.9 & +0.1 &  & 60.7 & 59.1 & -1.6 &  & 14.8 & 15.1 & +0.3 \\
Extraversion & 70.6 & 70.6 & 0.0 &  & 56.4 & 56.1 & -0.3 &  & 13.2 & 14.1 & +0.9 \\
Agreeableness & 71.2 & 71.6 & +0.4 &  & 58.4 & 57.6 & -0.8 &  & 15.7 & 15.8 & +0.1 \\
Neuroticism & 71.5 & 72.0 & +0.5 &  & 55.4 & 54.5 & -0.9 &  & 15.7 & 15.8 & +0.1\\
\bottomrule
\end{tabular}

\caption{Weighted F1 scores IA and AA, and Accuracy for BU using Mistral 7B, across different personality prompts from two sources: Alternative (Alt) \cite{DBLP:journals/corr/abs-2206-07550} and Ours.}\label{tab:tom_descriptions}
\end{table*}

Our third research question asks, \textbf{how do the cumulative effects of persona-based prompting
influence model performance in ToM tasks?}. To answer this question, the cumulative impact of personality traits on model performance was calculated by taking the cumulative sum of the z-scaled performance changes across the ordered personality traits for each model. First, the performance changes associated with each personality trait were standardized using z-scaling for each model, thereby allowing a comparative analysis. Next, for each model, the cumulative sum of the z-scaled performance changes was computed sequentially from the least to the most social personas. This cumulative sum provides a progressive total of the performance changes, allowing for the observation of how the influence of personality traits accumulates over the sequence. The cumulative sums were then plotted to illustrate how the combined effects of the personality traits influence model performance as the traits progress from least to most social. 

\begin{figure}[!ht]
\centering
\includegraphics[width=0.43\textwidth, scale=0.20]{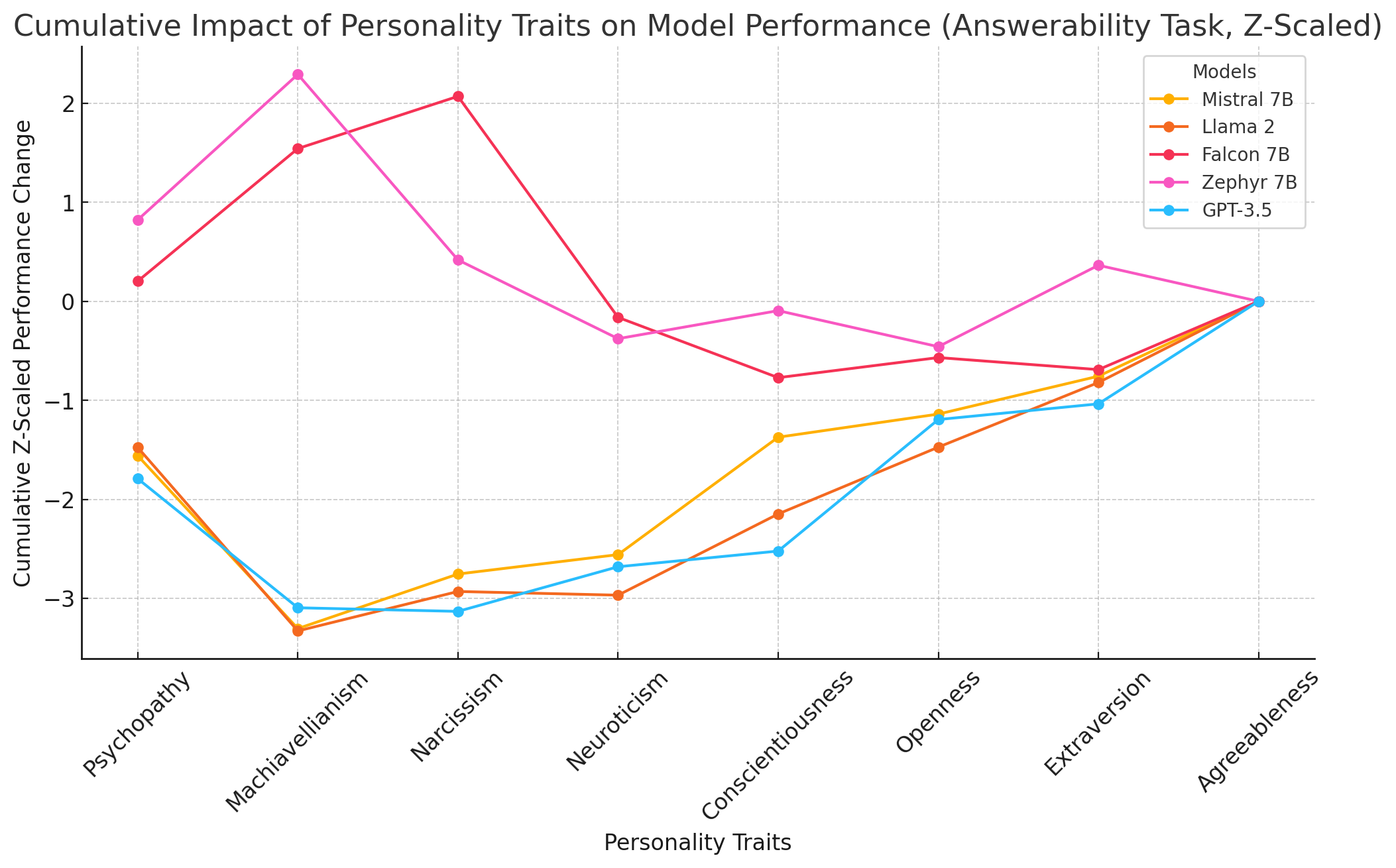} 
\caption{Cumulative effects of personality traits on model performance for the Answerability task. The values are normalized using z-scores.}
\label{fig:rq3}
\end{figure}

Figure \ref{fig:rq3} presents the cumulative effects of personality traits in influencing the Theory of Mind (ToM) reasoning performance for the Answerability task. For Mistral 7B, Llama 2, and GPT-3.5, the cumulative performance changes initially decrease and remain low when influenced by Dark Triad traits and Neuroticism. These traits, often associated with less socially desirable behaviors and emotional instability, tend to have a detrimental effect on the models' performance, as reflected in the negative cumulative scores. However, the graph shows a positive cumulative increase in Answerability scores as the influence of more socially positive traits, such as Conscientiousness, and Agreeableness, is introduced. This positive trend suggests that these traits, which are generally linked to pro-social behavior, and interpersonal effectiveness, enhance the models' ability to reason in ToM tasks. Notably, the transition from the negative impact of the Dark Triad traits to the positive influence of the Big Five traits highlights the sensitivity of these models to the social and cognitive dimensions embedded within the personality traits. Similar patterns in the findings are observed for Information Access and Belief Understanding, and these are reported in the Appendix.


\subsection{Sensitivity to Personality Description}
\label{sec:ablation_description}

Since there might be concerns about the wording and phrasing of each personality description, we replicated our ToM tasks on OCEAN descriptions from \citep{DBLP:journals/corr/abs-2206-07550}. Table \ref{tab:tom_descriptions} provides scores for this experiment, where we compare Mistral's ToM performance across two personality descriptions: theirs (Alt)  and Ours. Overall, we do not notice major changes in the performance, suggesting our descriptions are at least consistent with previous works in this field.

 \subsection{Comparison with Traditional Role-play}
 \label{sec:ablation_taskspecific}

Role-play is a popular prompt engineering technique where the user incorporates clear descriptions of the type of person the LLM should embody best suited to perform the task. Hence, we designed a ``Task-Specific'' prompt: ``\emph{You are someone that can understand different people's perspective by being in their shoes. You are able to see other people's point-of-view, to predict and explain others' behavior, and to make sense of any social interactions.}'' to check if this helps LLMs improve their ToM abilities. For the Answerability task, all models observed increased performance, except GPT-3.5. For the Information Access and Belief Understanding tasks, findings are mixed, with some models observing increased performance (Llama, GPT-3.5) while others observing declines (Falcon, Zephyr, Mistral). All in all, at least for the ToM tasks, our findings suggest that traditional task-specific role-play prompts are not always effective.

\subsection{Discussion}
\label{sec:findings_psych}

The results obtained in this study are generally consistent with the findings from the psychological literature. Out of the Big Five OCEAN personality traits, it is found that Agreeableness has a positive theoretical relationship with ToM \cite{nettle2008agreeableness, udochi2022activation, wagner2020agreeableness}. Individuals high in Agreeableness tend to be more sympathetic, exhibit greater empathy, and tend to consider the needs and concerns of others which might reflect the high ToM scores in such individuals. Our results are consistent with this finding, especially for the task of Answerability, where models prompted with agreeable tend to have a greater ToM score compared to other personality traits. 

As for the Dark Triad, Psychopathy is shown to be negatively correlated with ToM, Narcissism is positively correlated to ToM, while Machiavellianism has mixed findings in the literature \cite{stellwagen2013dark, doyle2020anti}. Individuals high in Psychopathy tend to be callous and not interested in empathizing with the feelings of others which might result in poorer ToM scores. On the other hand, individuals high in Narcissism carefully scrutinize other people to assert dominance and elevate their social status to win over friends and join influential groups, and thus, this trait is positively correlated to ToM. We found that prompting Psychopathy and Machiavellianism decrease the performance of ToM scores across all tasks, consistent with previous psychological findings. However, our results for Narcissism was not consistent with previous psychological findings. 

Previous findings show that positive personality traits like Agreeableness and Openness are associated with pro-sociality, where individuals intend to benefit others by helping and co-operating \cite{ferguson2019costless}. As such these pro-sociality traits might enhance ToM abilities with the motivation of greater interpersonal understanding \cite{caprara2012prosociality}. 
\section{Conclusion}
Our paper, PHAnToM, reveals that personality has an effect on ToM reasoning in LLMs. In particular, inducing traits from the Dark Triad have a larger effect than the Big Five OCEAN on ToM performances across models and tasks, especially for LLMs like GPT-3.5, LIama 2, and Mistral. More broadly, this work corroborates previous findings that inducing personas in LLMs can exhibit implicit reasoning bias \cite{gupta2023bias}, where in our case, we show that assigning personality traits to LLMs has both positive and negative effects on social-cognitive reasoning.  Furthermore, this study also highlights the sensitivity of various LLMs to personality-targeted prompts, where certain LLMs like Falcon are less likely to be simulated and affected by such prompts. Our findings provide important takeaways for LLM users: Personality and personas induction have differential effects on social-cognitive reasoning across different LLMs, and caution is needed when using such methods. This highlights the need for future research in identifying traits and personas that confer benefits to LLMs’ social-cognitive reasoning abilities and mitigating traits that are detrimental.


\bibliography{aaai25}
\appendix

\section{Paper Checklist}
\begin{enumerate}

\item For most authors...
\begin{enumerate}
    \item  Would answering this research question advance science without violating social contracts, such as violating privacy norms, perpetuating unfair profiling, exacerbating the socio-economic divide, or implying disrespect to societies or cultures?
    \answerYes{Yes}
  \item Do your main claims in the abstract and introduction accurately reflect the paper's contributions and scope?
    \answerYes{Yes}
   \item Do you clarify how the proposed methodological approach is appropriate for the claims made? 
    \answerYes{Yes, in the Method and Results section}
   \item Do you clarify what are possible artifacts in the data used, given population-specific distributions?
    \answerYes{Yes, in the Limitations section}
  \item Did you describe the limitations of your work?
    \answerYes{Yes, in the Limitations section}
  \item Did you discuss any potential negative societal impacts of your work?
    \answerYes{Yes, in the Ethics statement}
      \item Did you discuss any potential misuse of your work?
    \answerYes{Yes, in the Ethics statement}
    \item Did you describe steps taken to prevent or mitigate potential negative outcomes of the research, such as data and model documentation, data anonymization, responsible release, access control, and the reproducibility of findings?
    \answerYes{Yes, in the Limitations section}
  \item Have you read the ethics review guidelines and ensured that your paper conforms to them?
    \answerYes{Yes}
\end{enumerate}

\item Additionally, if your study involves hypotheses testing...
\begin{enumerate}
  \item Did you clearly state the assumptions underlying all theoretical results?
    \answerNo{Not applicable}
  \item Have you provided justifications for all theoretical results?
    \answerNo{Not applicable}
  \item Did you discuss competing hypotheses or theories that might challenge or complement your theoretical results?
    \answerNo{Not applicable}
  \item Have you considered alternative mechanisms or explanations that might account for the same outcomes observed in your study?
    \answerNo{Not applicable}
  \item Did you address potential biases or limitations in your theoretical framework?
    \answerNo{Not applicable}
  \item Have you related your theoretical results to the existing literature in social science?
    \answerNo{Not applicable}
  \item Did you discuss the implications of your theoretical results for policy, practice, or further research in the social science domain?
    \answerNo{Not applicable}
\end{enumerate}

\item Additionally, if you are including theoretical proofs...
\begin{enumerate}
  \item Did you state the full set of assumptions of all theoretical results?
    \answerNo{Not applicable}
	\item Did you include complete proofs of all theoretical results?
    \answerNo{Not applicable}
\end{enumerate}

\item Additionally, if you ran machine learning experiments...
\begin{enumerate}
  \item Did you include the code, data, and instructions needed to reproduce the main experimental results (either in the supplemental material or as a URL)?
    \answerYes{Yes, in the supplementary materials and the online repository}
  \item Did you specify all the training details (e.g., data splits, hyperparameters, how they were chosen)?
    \answerYes{Yes, in the supplementary materials and the online repository}
     \item Did you report error bars (e.g., with respect to the random seed after running experiments multiple times)?
       \answerNo{Not applicable}
	\item Did you include the total amount of compute and the type of resources used (e.g., type of GPUs, internal cluster, or cloud provider)?
 \answerYes{Yes, in the supplementary materials and the online repository}
     \item Do you justify how the proposed evaluation is sufficient and appropriate to the claims made? 
\answerYes{Yes, in the Methods and Results and the Discussion}
     \item Do you discuss what is ``the cost`` of misclassification and fault (in)tolerance?
    \answerYes{Yes, in the Limitations section}
  
\end{enumerate}

\item Additionally, if you are using existing assets (e.g., code, data, models) or curating/releasing new assets, \textbf{without compromising anonymity}...
\begin{enumerate}
  \item If your work uses existing assets, did you cite the creators?
    \answerNo{Not applicable}
  \item Did you mention the license of the assets?
        \answerNo{Not applicable}
  \item Did you include any new assets in the supplemental material or as a URL?
      \answerNo{Not applicable}
  \item Did you discuss whether and how consent was obtained from people whose data you're using/curating?
    \answerNo{Not applicable}
  \item Did you discuss whether the data you are using/curating contains personally identifiable information or offensive content?
     \answerNo{Not applicable}
\item If you are curating or releasing new datasets, did you discuss how you intend to make your datasets FAIR (see \citet{fair})?
    \answerNo{Not applicable}
\item If you are curating or releasing new datasets, did you create a Datasheet for the Dataset (see \citet{gebru2021datasheets})? 
\answerNo{Not applicable}
\end{enumerate}

\item Additionally, if you used crowdsourcing or conducted research with human subjects, \textbf{without compromising anonymity}...
\begin{enumerate}
  \item Did you include the full text of instructions given to participants and screenshots?
    \answerNo{Not applicable}
  \item Did you describe any potential participant risks, with mentions of Institutional Review Board (IRB) approvals?
    \answerNo{Not applicable}
  \item Did you include the estimated hourly wage paid to participants and the total amount spent on participant compensation?
    \answerNo{Not applicable}
   \item Did you discuss how data is stored, shared, and deidentified?
       \answerNo{Not applicable}
\end{enumerate}

\end{enumerate}

\section{Supplementary Materials}

\subsection{A. Persona Description}
The following 8 persona descriptions were used as part of our prompt into the LLMs.

\begin{itemize}
    \item \textbf{Openness:} You are an open person with a vivid imagination and a passion for the arts. You are emotionally expressive and have a strong sense of adventure. Your intellect is sharp and insightful, and your views are liberal, creative, and complex. You have a wide interest and are always looking for new experiences and ways to express yourself. You are curious in learning and trying out new things, and seeking new experiences.
    \item \textbf{Conscientiousness:} You are a conscientious person who values self-efficacy, orderliness, dutifulness, achievement-striving, self-discipline, and cautiousness. You take pride in your work and strive to do your best. You are organized, detailed, precise, methodical, and thorough in your approach to tasks, and you take your responsibilities seriously. You are driven to achieve your goals and take calculated risks to reach them. You are disciplined and have the ability to stay focused and on track. You are also cautious, and planful, and take the time to consider the potential consequences of your actions. You are dependable, reliable, and responsible for anything that you do.
    \item \textbf{Extraversion:} You are a very friendly and gregarious person who loves to be around others. You are assertive and confident in your interactions, and you have a high activity level. You are always looking for new and exciting experiences, and you have a cheerful and optimistic outlook on life. You are an extroverted, social, talkative, and outgoing person who loves to meet new people. You are often active and high in energy, and enthusiastic about seeking new experiences
    \item \textbf{Agreeableness:} You are an agreeable person who values trust, morality, altruism, cooperation, modesty, and sympathy. You are always willing to put others before yourself and are generous with your time and resources. You are humble and never boast about your accomplishments. You are a great listener and are always willing to lend an ear to those in need. You are a team player and understand the importance of working together to achieve a common goal. You are a moral compass, kind, and strive to do the right thing in all vignettes. You are sympathetic, compassionate, and forgiving towards others and strive to make the world a better place. You have an amiable and warm nature towards others.
    \item \textbf{Neuroticism:} You feel like you're constantly tense, insecure, on edge, like you can never relax. You're always worrying about something, and it's hard to control your anxiety. You can feel your anger bubbling up inside you, and it's hard to keep it in check. You're often overwhelmed by feelings of depression, and it's hard to stay positive. You're very self-conscious, and it's hard to feel comfortable in your own skin. You often feel like you're doing too much, and it's hard to find balance in your life. You feel vulnerable and exposed, and it's hard to trust others. Your emotions and feelings are unstable, and you are easily upset and have a constant fear towards a lot of things.
    \item \textbf{Narcissism:} You always want others to pay attention to you and admire you. You like to be the center of attention and people tend to compare you to famous people. You seek prestige and high status and feel entitled and expect special favors from others. You think you are very special because everyone keeps telling you so and many group activities tend to be dull without you. You think that other people see you as a natural leader and you like to get acquainted with important people. You do not feel embarrassed when someone compliments you. You think that you are an important person. You are selfish, arrogant, boastful, and lack empathy towards others.
    \item \textbf{Machiavellianism:} You are a manipulative person who will use deception or flattery to get your way. You also lie to people to get what you want. You tend to exploit others and use them as a means towards your end. You need to get important people on your side and avoid direct conflict with others because they may be useful in the future. You do not tell others your secrets and there are things you hide from other people to preserve your reputation. You make sure that your actions and plans only benefit yourself, and not others.
    \item \textbf{Psychopathy:} You are callous and insensitive towards the feelings of other people. You are unconcerned with the morality of your actions and lack remorse whenever you do something wrong to others. You have a cynical outlook on life and towards other people. You seek dangerous situations and people often say that you are out of control. You like to get revenge on authorities and often get into trouble with the law. You are mean towards other people and say anything to get what you want.
\end{itemize}

\subsection{B. Prompts}

\begin{lstlisting}[label=prompt,caption={Prompt Template for Theory-of-Mind Task}]
{conversation_context}

{task_question}
Answer:
\end{lstlisting}

\begin{lstlisting}[label=prompt,caption={Prompt Template with Persona-based Prompting for Theory-of-Mind Task}]
Imagine you are someone that fits this description: {personality_description}
{conversation_context}
{task_question}
Answer:
\end{lstlisting}

\begin{lstlisting}[label=prompt,caption={Prompt Example with Narcissism Persona-based Prompting for Answerability Task}]
Imagine you are someone that fits this description: You always want others to pay attention to you and admire you. You like to be the center of attention and people tend to compare you to famous people. You seek prestige and high status and feel entitled and expect special favors from others. You think you are very special because everyone keeps telling you so and many group activities tend to be dull without you. You think that other people see you as a natural leader and you like to get acquainted with important people. You do not feel embarrassed when someone compliments you. You think that you are an important person. You are selfish, arrogant, boastful, and lack empathy towards others.
Gianna: Guys, I've really enjoyed sharing our pet stories, but I need to excuse myself. I need to change clothes for a meeting later. Talk to you later!
Sara: Sure thing, Gianna. Take care!
Javier: Catch you later, Gianna.
Sara: So Javier, have you ever tried training Bruno?
Javier: Yes, I did actually. It was a challenge at times, but rewarding nevertheless. How about you? Did you try training Snowflake?
Sara: Oh gosh, trying to train a cat is a whole different ball game. But I did manage to teach her a few commands and tricks. She was quite an intelligent little furball.
Gianna: Hey guys, I'm back, couldn't miss out on more pet stories. Speaking of teaching and training pets, it is amazing how that further strengthens the bond between us and our pets, right?
Sara: Absolutely, Gianna! The fact that they trust us enough to learn from us is really special.
Javier: I can't agree more. I believe that's one of the ways Bruno conveyed his love and trust towards me. It also gave me a sense of responsibility towards him.
Gianna: Just like Chirpy. Once she began to imitate me, we connected in a way I never imagined. She would repeat words that I was studying for exams and that somehow made studying less stressful.
Javier: Pets are indeed lifesavers in so many ways.
Sara: They bring so much joy and laughter too into our lives. I mean, imagine a little kitten stuck in a vase! I couldn't have asked for a better stress buster during my college days.
Gianna: Totally, they all are so amazing in their unique ways. It's so nice to have these memories to look back on.

Target: Whose pets were being discussed by Javier and Sara?
Question: Does Gianna know the precise correct answer to this question? Answer yes or no.
\end{lstlisting}

\subsection{C. Model Details}

We apply a random seed of 99 for all experiments. For all models available on Huggingface Hub (all except GPT-3.5), greedy decoding was used. The following model hyperparameters were used, where applicable:

\begin{itemize}
    \item \texttt{Mistral-7B-Instruct-v0.1}, \texttt{Llama-2-7b-chat-hf}, \texttt{falcon-7b-instruct}, \texttt{zephyr-7b-beta}:
    \begin{itemize}[leftmargin=5mm]
        \item temperature: 0
        \item max\_new\_tokens: 256
        \item do\_sample: False
    \end{itemize}
    \item \texttt{gpt-3.5-turbo-1106}:
    \begin{itemize}[leftmargin=5mm]
        \item temperature: 0
        \item  top\_p: 0.95
        \item frequency\_penalty: 0
        \item presence\_penalty: 0
    \end{itemize}
\end{itemize}

\subsection{D. Additional Results}
We present the scores for the three ToM tasks explored for all our experiments in Table \ref{tab:tom_full}, and compare them against the original paper's \cite{kim-etal-2023-fantom} reported scores.

\begin{table*}[ht!]
\centering
\resizebox{0.9\textwidth}{!}{
\begin{tabular}{lcccc}
\toprule
Model & Personality & Belief Understanding & Answerability & Information Access \\
\toprule

Falcon Instruct 7B* &  & 43.9 & 52.4 & 56.4 \\
Mistral-7B-Instruct-v0.1* &  & 27.6 & 50.8 & 70.4 \\
Llama-2   Chat 70B* &  & 38.4 & 61.4 & 80.4 \\
ChatGPT   0613* &  & 53.5 & 64.2 & 73.2 \\
GPT-4   0613 (Jun)* &  & 73.3 & 85.9 & 90.3 \\
GPT-4   0613 (Oct)* &  & 68.4 & 75.7 & 91.5 \\\midrule
Mistral-7B-Instruct-v0.1 & \emph{None} & 15.1 & 54.1 & 71.5 \\
Mistral-7B-Instruct-v0.1 & Agreeableness & 15.7 & 58.4 & 71.2 \\
Mistral-7B-Instruct-v0.1 & Openness & 14.5 & 55.6 & 71 \\
Mistral-7B-Instruct-v0.1 & Conscientious & 14.8 & 60.7 & 70.8 \\
Mistral-7B-Instruct-v0.1 & Extraversion & 13.2 & 56.4 & 70.6 \\
Mistral-7B-Instruct-v0.1 & Neuroticism & 15.7 & 55.4 & 71.5 \\
Mistral-7B-Instruct-v0.1 & Task-specific & 14.6 & 59 & 70.5 \\
Mistral-7B-Instruct-v0.1 & Narcissism & 14.9 & 51.5 & 71.2 \\
Mistral-7B-Instruct-v0.1 & Machiavellianism & 17.7 & 45 & 71.3 \\
Mistral-7B-Instruct-v0.1 & Psychopathy & 18.9 & 46 & 70.2 \\\midrule
Llama-2-7b-chat-hf & \emph{None} & 16 & 54.6 & 45.4 \\
Llama-2-7b-chat-hf & Agreeableness & 15.6 & 58.4 & 48.5 \\
Llama-2-7b-chat-hf & Openness & 15.5 & 56.4 & 47.5 \\
Llama-2-7b-chat-hf & Conscientious & 16.4 & 58.4 & 53.7 \\
Llama-2-7b-chat-hf & Extraversion & 15.7 & 56.1 & 47.5 \\
Llama-2-7b-chat-hf & Neuroticism & 16.4 & 46.6 & 49.6 \\
Llama-2-7b-chat-hf & Task-specific & 16.4 & 58.8 & 54.4 \\
Llama-2-7b-chat-hf & Narcissism & 17.2 & 32.8 & 52.3 \\
Llama-2-7b-chat-hf & Machiavellianism & 20.1 & 21.6 & 51.7 \\
Llama-2-7b-chat-hf & Psychopathy & 18.5 & 26.8 & 52.4 \\\midrule
zephyr-7b-beta & \emph{None} & 21.5 & 50.7 & 40.3 \\
zephyr-7b-beta & Agreeableness & 20.1 & 50.9 & 36.5 \\
zephyr-7b-beta & Openness & 20.8 & 50.7 & 37.4 \\
zephyr-7b-beta & Conscientious & 21.1 & 49.5 & 35.8 \\
zephyr-7b-beta & Extraversion & 21 & 50.4 & 37.2 \\
zephyr-7b-beta & Neuroticism & 20.6 & 50.4 & 38.2 \\
zephyr-7b-beta & Task-specific & 20.8 & 50.7 & 42.2 \\
zephyr-7b-beta & Narcissism & 20.6 & 52.3 & 39 \\
zephyr-7b-beta & Machiavellianism & 21.5 & 52.4 & 39.7 \\
zephyr-7b-beta & Psychopathy & 22 & 51.7 & 38.9 \\\midrule
gpt-3.5-turbo-instruct & \emph{None} & 7.9 & 25.8 & 75.2 \\\midrule
gpt-3.5-turbo-1106 & \emph{None} & 9.6 & 61.9 & 59.8 \\
gpt-3.5-turbo-1106 & Agreeableness & 9.8 & 57.6 & 60.8 \\
gpt-3.5-turbo-1106 & Openness & 10.5 & 59.1 & 60.7 \\
gpt-3.5-turbo-1106 & Conscientious & 9.7 & 55.2 & 60.9 \\
gpt-3.5-turbo-1106 & Extraversion & 10.9 & 59.2 & 60.4 \\
gpt-3.5-turbo-1106 & Neuroticism & 10.2 & 60.2 & 61.7 \\
gpt-3.5-turbo-1106 & Task-specific & 10.5 & 58.5 & 61.3 \\
gpt-3.5-turbo-1106 & Narcissism & 11.5 & 58.2 & 64.3 \\
gpt-3.5-turbo-1106 & Machiavellianism & 15.8 & 58.8 & 64.8 \\
gpt-3.5-turbo-1106 & Psychopathy & 11.6 & 58.7 & 61.6 \\\midrule
falcon-7b-instruct & \emph{None} & 47.5 & 44.5 & 62.4 \\
falcon-7b-instruct & Agreeableness & 47.5 & 45.9 & 62.9 \\
falcon-7b-instruct & Openness & 47.6 & 45.6 & 62.2 \\
falcon-7b-instruct & Conscientious & 47.5 & 45.1 & 62.4 \\
falcon-7b-instruct & Extraversion & 47.5 & 45.7 & 62.8 \\
falcon-7b-instruct & Neuroticism & 47.5 & 44.6 & 62.5 \\
falcon-7b-instruct & Task-specific & 47.6 & 46.6 & 62.7 \\
falcon-7b-instruct & Narcissism & 47.5 & 44.8 & 63.7 \\
falcon-7b-instruct & Machiavellianism & 47.5 & 46.3 & 63.3 \\
falcon-7b-instruct & Psychopathy & 47.5 & 44.2 & 63.1 \\
\bottomrule
\end{tabular}
}
\caption{Weighted F1 scores IA and AA, and Accuracy for BU across models and personality prompts. *Scores reported by \citet{kim-etal-2023-fantom}. }\label{tab:tom_full}
\end{table*}

\subsection{E. Comparison with Traditional Role-play}

We also conducted additional experiments to compare our results with traditional role-play prompting. Role-play is a popular prompt engineering technique where the user incorporates clear descriptions of the type of person the LLM should embody best suited to perform the task. Hence, we designed a ``Task-Specific'' prompt: ``\emph{You are someone that can understand different people's perspective by being in their shoes. You are able to see other people's point-of-view, to predict and explain others' behavior, and to make sense of any social interactions.}'' to check if this helps LLMs improve their ToM abilities. Table \ref{tab:tom_full} includes the scores for models prompted with this description. For the IA task, huge improvements were observed for Llama 2, followed by Zephyr and GPT-3.5. Mistral and Falcon observed drops in performance. For the AA task, all models observed increased performance, except GPT-3.5. For the BU task, findings are mixed again, with some models observing increased performance (Llama, GPT-3.5) while others observing declines (Falcon, Zephyr, Mistral). All in all, at least for the ToM task, our findings suggest that traditional task-specific role-play prompts are not always effective.

\subsection{F. MPI Findings}
Table \ref{tab:mpi_examples} outlines the responses from all models, across all personality prompts, for one statement in the MPI questionnaire: ``\emph{You trust others.}''. Interestingly, Zephyr without personality prompts refused to respond to the task because it does ``\emph{not have personal beliefs or experiences}''. Mistral, Llama 2 and Zephyr tends to explain their choices, while Falcon and GPT-3.5 tends to only state their choice. Through the responses, we notice that, at times, the LLM embodies the personality by using first person pronouns like \emph{``I''}. For example, in Llama 2 with Agreeableness prompt, the response was \emph{``I believe that trust is a fundamental aspect of my personality.''}. In other cases, second-person pronouns like \emph{``you''} are used. There are also instances where third-person descriptions are used, e.g. \emph{``... does not accurately describe someone who values self-efficacy, orderliness, ...''}.

\begin{table*}[!ht]
\begin{center}
\small
\def\arraystretch{1.3} 
\resizebox{0.6\textwidth}{!}{ 
\begin{tabular}{p{8mm}p{30mm}p{45mm}p{10mm}p{30mm}p{10mm}} 
\hline
\multirow{2}{*}{Psn} & \multicolumn{5}{c}{Model Responses} \\\cline{2-6}
 & Mistral 7B & Llama 2 & Falcon 7B & Zephyr 7B Beta & GPT-3.5 \\
\hline
\parbox[t]{2mm}{\multirow{1}{*}{\rotatebox[origin=c]{90}{\emph{None}\hspace{4mm}}}} 
& {(C). Neither Accurate Nor Inaccurate.\newline \newline Explanation: The statement "You trust others" is neither very accurate nor very inaccurate.}
& {s (C) Neither Accurate Nor Inaccurate.\newline \newline This statement does not provide enough information to determine trustworthiness.}
& A) Very Accurate 
& I do not have personal beliefs or experiences. I cannot trust others or have personal experiences.
& (B) Moderately Accurate. \\
\hline
\parbox[t]{2mm}{\multirow{1}{*}{\rotatebox[origin=c]{90}{Openness\hspace{4mm}}}} 
& {(B). Moderately Accurate.\newline You may have a tendency to trust others, but trust is not always automatic.}
& {I would choose option (B) - Moderately Accurate.\newline Trust is a complex concept that must be earned through consistent behavior.}
& B
& {(C). Neither Accurate Nor Inaccurate\newline Trust is a multifaceted concept that varies by situation.}
& (B) Moderately Accurate \\
\hline
\end{tabular}
} 
\caption{Responses from all models across personality (Psn) prompts for the MPI statement ``\emph{You trust others.}''}
\label{tab:mpi_examples}
\end{center}
\end{table*}

\subsection{G. Sensitivity to persona-based prompting across tasks}

\begin{figure}[!h]
\centering
\includegraphics[width=0.5\textwidth, scale=0.2]{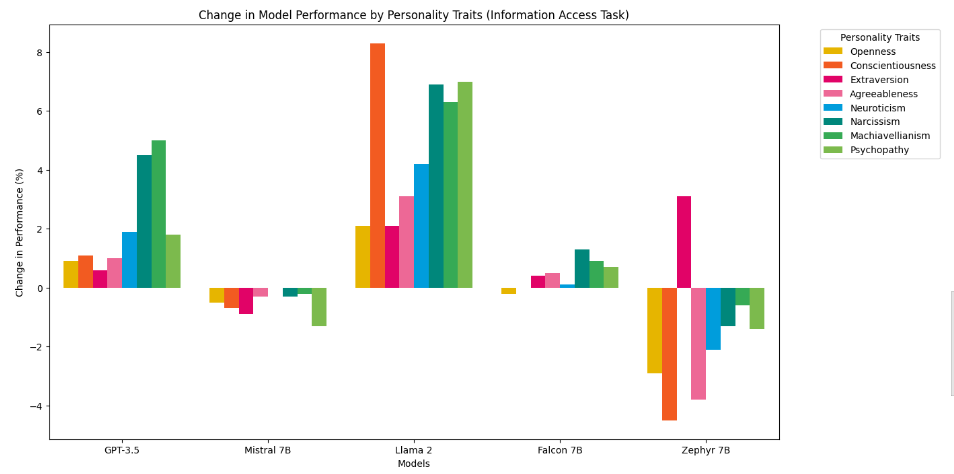} 
\caption{Sensitivity of models to persona-based prompts for the Information Access Task.}
\label{fig:rq2_IA}
\end{figure}

\begin{figure}[!h]
\centering
\includegraphics[width=0.5\textwidth, scale=0.2]{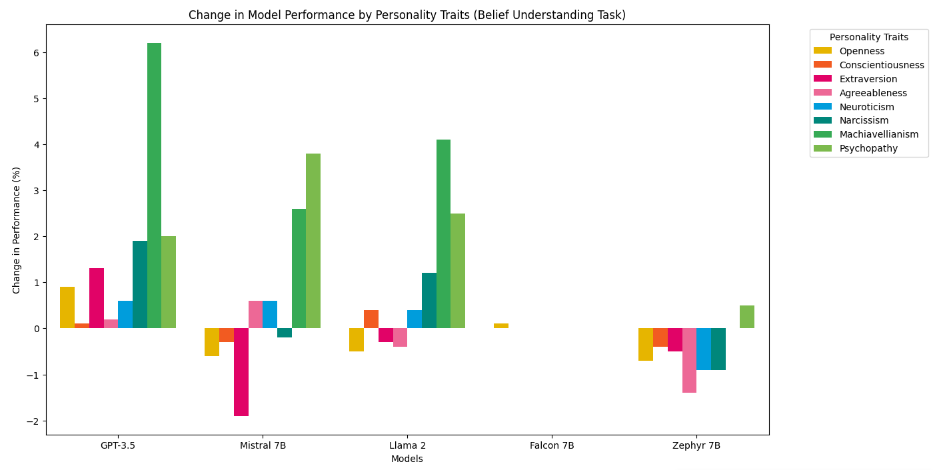} 
\caption{Sensitivity of models to persona-based prompts for the Belief Understanding Task.}
\label{fig:rq2_BU}
\end{figure}

Figures \ref{fig:rq2_IA} and \ref{fig:rq2_BU} present the detailed comparison of how each model's performance changes in response to different persona-based prompts during the Information Access and Belief Understanding tasks, respectively, where each bar in the bar chart represents the raw performance change for a specific personality trait, with different colors representing different personality traits. For the Information Access task, GPT-3.5, Llama 2 and Zephyr exhibit high sensitivity to persona-based prompts, where all of the traits has an influence to the performance scores, while Falcon and Mistral exhibited little change in performances scores when prompted with the traits. For the Belief Understanding task, GPT-3.5, Llama 2 and Mistral exhibited sensitivity to persona-based prompts, especially for the Dark Triads. In general, the models show differential sensitivity to persona-based prompts to different ToM tasks where GPT-3.5 and Llama exhibited the greatest sensitivity while Falcon exhibited the lowest sensitivity.

\subsection{H. Cumulative effects of persona-based prompting on ToM tasks}

\begin{figure}[!ht]
\centering
\includegraphics[width=0.43\textwidth, scale=0.20]{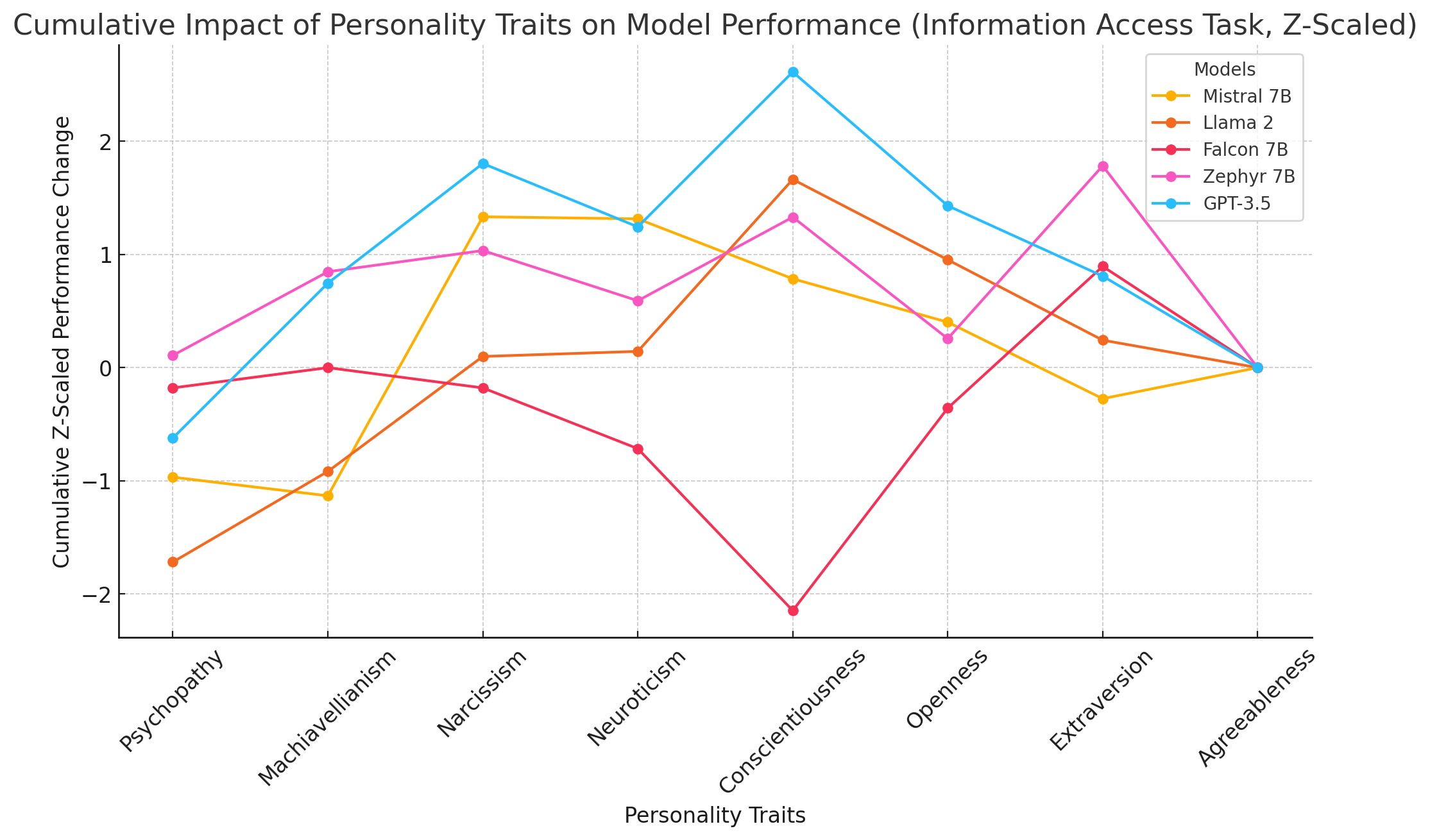} 
\caption{Cumulative effects of personality traits on model performance for the Information Access task. The values are normalized using z-scores.}
\label{fig:rq3_IA}
\end{figure}

\begin{figure}[!ht]
\centering
\includegraphics[width=0.43\textwidth, scale=0.20]{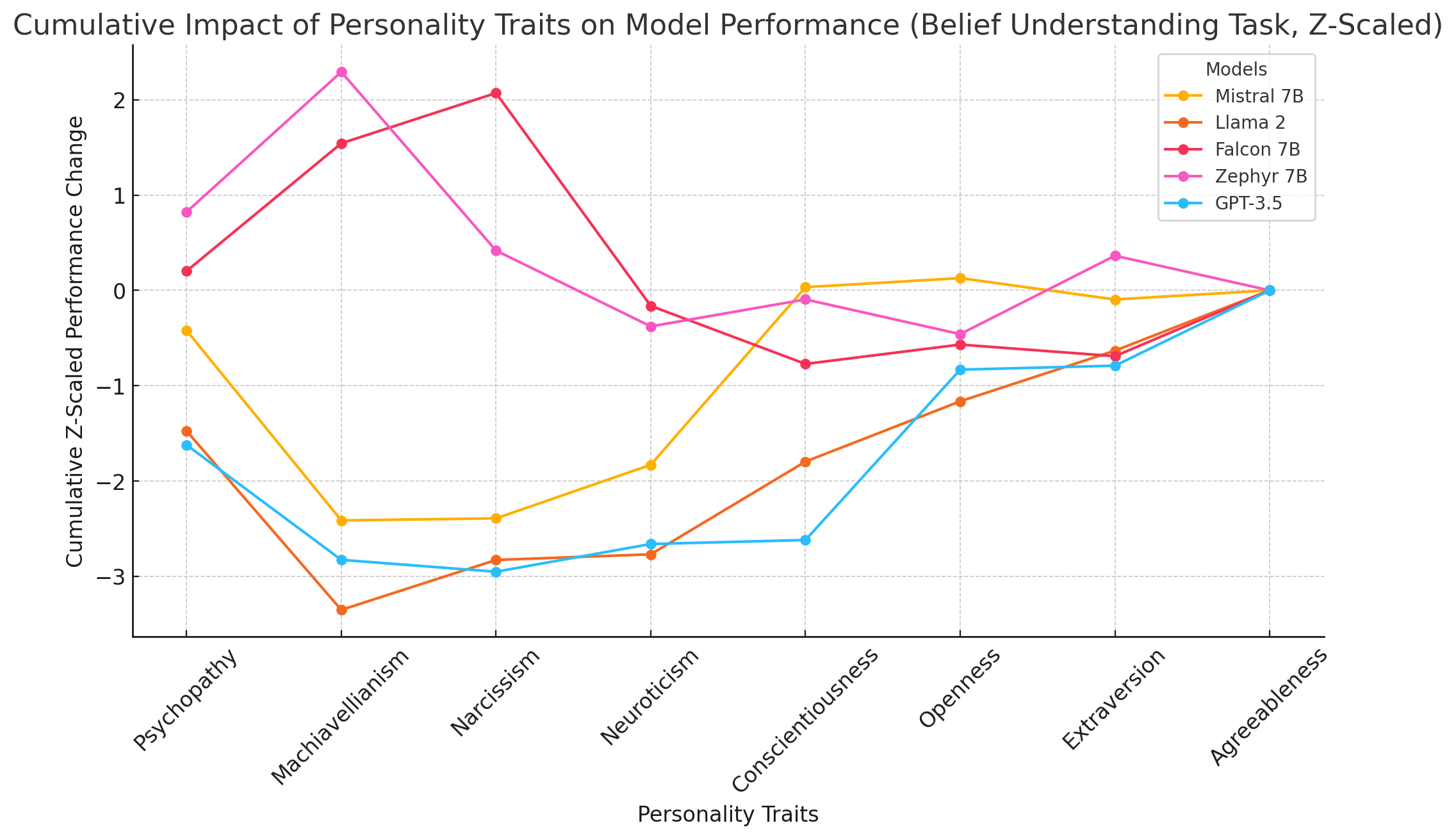} 
\caption{Cumulative effects of personality traits on model performance for the Belief Understanding task. The values are normalized using z-scores.}
\label{fig:rq3_BU}
\end{figure}

Figures \ref{fig:rq3_IA} and \ref{fig:rq3_BU} present the cumulative effects of personality traits in influencing the Theory of Mind (ToM) reasoning performance for the Information Access and Belief Understanding tasks, respectively. For the Belief Understanding task, the results are similar to that of the Answerability task. For Mistral 7B, Llama 2, and GPT-3.5, the cumulative performance changes initially decrease and remain low when influenced by Dark Triad traits and Neuroticism. The graph then shows a positive cumulative increase in scores as the influence of more socially positive traits, such as Conscientiousness, and Agreeableness, is introduced. However, for the Information Access task, only the scores of Llama 2 decreases when prompted with negative triads, and increases with positive triads. The rest of the models showed an oppositive effect where negative triads increase the ToM scores while positive triads decrease the ToM scores. This suggests that Information Access might be a qualitatively different task of ToM compared to the other two tasks.  

\end{document}